# Integrative Windowing

**Johannes Fürnkranz** JUFFI@CS.CMU.EDU
*School of Computer Science*
*Carnegie Mellon University*
*Pittsburgh, PA 15213*

## Abstract

In this paper we re-investigate windowing for rule learning algorithms. We show that, contrary to previous results for decision tree learning, windowing can in fact achieve significant run-time gains in noise-free domains and explain the different behavior of rule learning algorithms by the fact that they learn each rule independently. The main contribution of this paper is integrative windowing, a new type of algorithm that further exploits this property by integrating good rules into the final theory right after they have been discovered. Thus it avoids re-learning these rules in subsequent iterations of the windowing process. Experimental evidence in a variety of noise-free domains shows that integrative windowing can in fact achieve substantial run-time gains. Furthermore, we discuss the problem of noise in windowing and present an algorithm that is able to achieve run-time gains in a set of experiments in a simple domain with artificial noise.

## 1. Introduction

Windowing is a sub-sampling technique proposed by Quinlan (1983) for the ID3 decision tree learning algorithm. Its goal is to reduce the complexity of a learning problem by identifying an appropriate subset of the original data, from which a theory of sufficient quality can be learned. For this purpose, it maintains a subset of the available data, the so-called *window*, which is used as the training set for the learning algorithm. The window is initialized with a small random sample of the available data, and the learning algorithm induces a theory from this sample. This theory is then tested on the remaining examples. If the quality of the learned theory is not sufficient, the window is adjusted, usually by adding more examples from the training data, and a new theory is learned. This process is repeated until a theory of sufficient quality has been found.

There are at least three motivations for studying windowing techniques:

**Memory Limitations:** Almost all learning algorithms still require to have all training examples and all background knowledge in main memory. Although memory has become cheap and the capacity of the main memory of the available hardware platforms is increasing rapidly, there certainly are datasets too big to fit into the main memory of conventional computer systems.

**Efficiency Gain:** Learning time usually increases (most often super-linearly) with the complexity of a learning problem. Reducing this complexity may be necessary to make a learning problem tractable.

```
procedure Windowing(Examples,InitSize,MaxIncSize)

Window = RandomSample(Examples,InitSize)
Test = Examples \ Window
repeat
    Theory = Induce(Window)
    NewWindow = ∅
    OldTest = ∅
    for Example ∈ Test
        Test = Test \ Example
        if Classify(Theory,Example) ≠ Class(Example)
            NewWindow = NewWindow ∪ Example
        else
            OldTest = OldTest ∪ Example
        if |NewWindow| = MaxIncSize
            exit for
    Test = Append(Test,OldTest)
    Window = Window ∪ NewWindow
until NewWindow = ∅
return(Theory)
```

Figure 1: The basic windowing algorithm

**Accuracy Gain:** It has been observed that windowing may also lead to an increase in predictive accuracy. A possible explanation for this phenomenon is that learning from a subset of examples may often result in a less over-fitting theory.

In this paper, our major concern is the appropriateness of windowing techniques for increasing the efficiency of inductive rule learning algorithms, such as those using the popular separate-and-conquer (or covering) learning strategy (Michalski, 1969; Clark & Niblett, 1989; Quinlan, 1990; Fürnkranz, 1998). We will argue that windowing is more suitable for these algorithms than for divide-and-conquer decision-tree learning (Quinlan, 1983, 1993) (section 3. Thereafter, we will introduce *integrative windowing*, a technique that exploits the fact that rule learning algorithms learn each rule independently. We will show that this method allows to significantly improve the performance of windowing by integrating good rules learned from different iterations of the basic windowing procedure into the final theory (section 4). While we have primarily worked with noise-free domains, section 5 will discuss the problem of noise in windowing as well as some preliminary work that shows how windowing techniques can be adapted for noisy domains. Parts of this work have previously appeared as (Fürnkranz, 1997c, 1997d, 1997a).

## 2. A Brief History of Windowing

Windowing dates back to early versions of the ID3 decision tree learning algorithm, where it was devised as an automated teaching procedure that allowed a preliminary version of ID3 to discover complete and consistent descriptions of various problems in a KRKN chess endgame (Quinlan, 1979). Figure 1 shows the basic windowing algorithm as described in Quinlan's subsequent seminal paper on ID3 (Quinlan, 1983). It starts by picking a random sample of a user-settable size *InitSize* from the total set of *Examples*. These examples are used for inducing a theory with a given learning algorithm. This theory is then tested on the remaining examples and all misclassified examples are removed from the test set and added to the window of the next iteration. In order to keep the size of the training set small, another parameter, *MaxIncSize*, controls the maximum number of examples that can be added to the training set in one iteration. If this number is reached, no further examples are tested and the next theory is learned from the new training set. To make sure that all examples are tested in the first few iterations, the examples that have already been tested (*OldTest*) are Appended to the examples that have not yet been used, so that testing will start with new examples in the next iteration.[1]

Quinlan (1983) argued that windowing is necessary for ID3 to tackle very large classification problems. Nevertheless, windowing has not played a major role in machine learning research. One reason for this certainly is the rapid development of computer hardware, which made the motivation for windowing seem less compelling. However, a good deal of this lack of interest can be attributed to an empirical study (Wirth & Catlett, 1988), in which the authors studied windowing with ID3 in various domains and concluded that it cannot be recommended as a procedure for improving efficiency. The best results were achieved in noise-free domains, such as the *Mushroom* domain, where windowing was able to perform on the same level as simple ID3, while its performance in noisy domains was considerably worse.

Despite the discouraging experimental evidence of Wirth and Catlett (1988), Quinlan (1993) implemented a new version of windowing into the C4.5 learning algorithm. It differs from the simple windowing version originally proposed for ID3 (Quinlan, 1983) in several ways:

- While selecting examples, it takes care to make the class distribution as uniform as possible. This can lead to accuracy gains in domains with skewed distributions (Catlett, 1991b).
- It includes at least half of the misclassified examples into next window, which supposedly guarantees faster convergence (fewer iterations) in noisy domains.
- It can stop before all examples are correctly classified, if it appears that no further gains in accuracy are possible.
- C4.5's `-t` parameter, which invokes windowing, allows it to perform multiple runs of windowing and to select the best tree.

Nevertheless, windowing is arguably one of C4.5's least frequently used options.

Recent work in the areas of *Knowledge Discovery in Databases* (Kivinen & Mannila, 1994; Toivonen, 1996) and *Intelligent Information Retrieval* (Lewis & Catlett, 1994; Yang, 1996) has re-emphasized the importance of sub-sampling procedures for reducing both learning time and memory requirements. Thus the interest in windowing techniques has revived as well. We discuss some of the more recent approaches in section 6.

1. Quinlan does not explicitly specify how this case should be handled, but we think it makes sense that way.

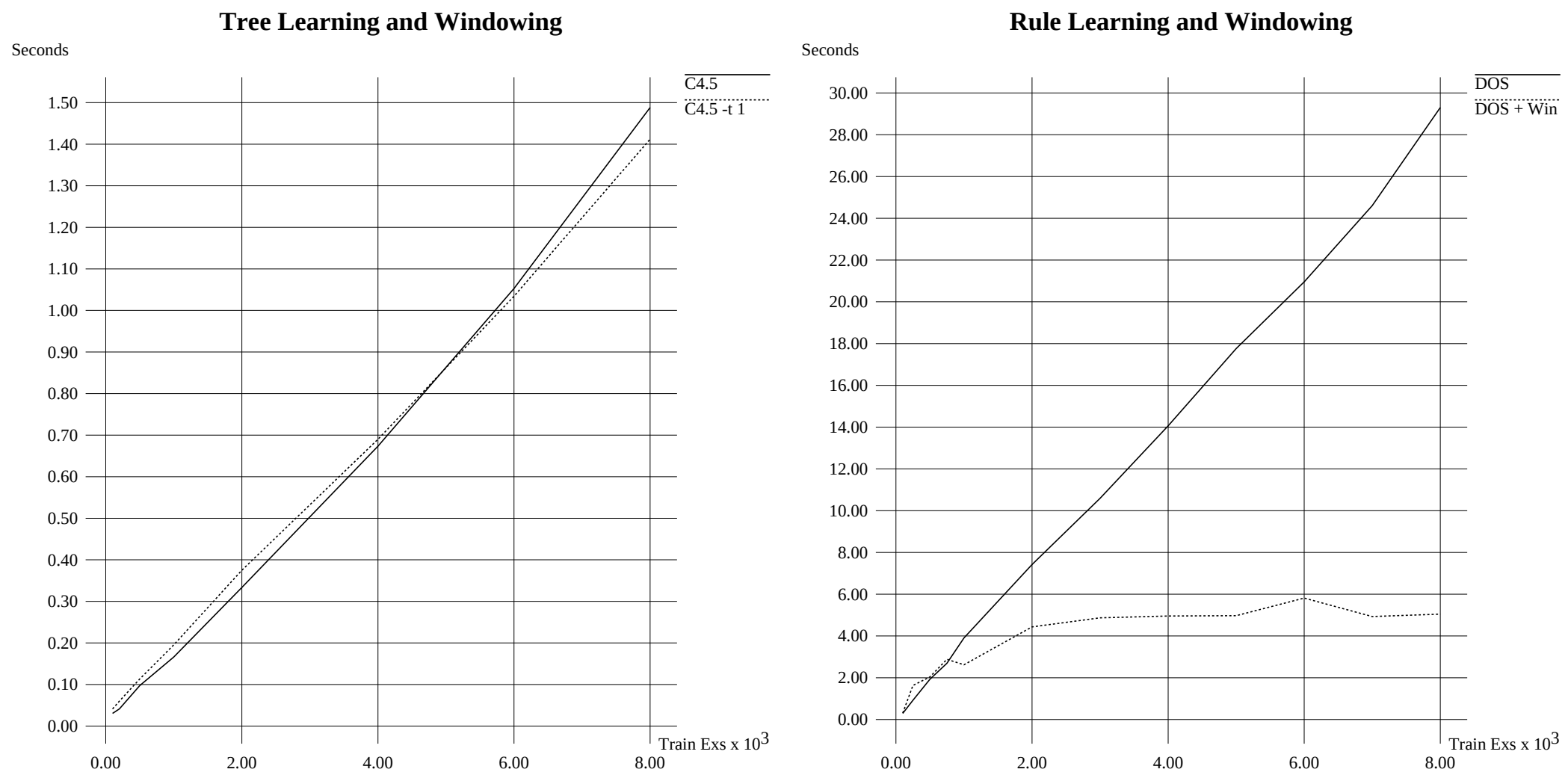

Figure 2: Results for windowing with the decision tree learner `c4.5` and a rule learner in the *Mushroom* domain.

## 3. A Closer Look at Windowing

### 3.1 A Motivating Example

The motivation for our study came from a brief experiment in which we compared windowing with a decision tree algorithm to windowing with a rule learning algorithm in the noise-free *Mushroom* domain. This domain contains 8124 examples represented by 22 symbolic attributes. The task is to discriminate between poisonous and edible mushrooms. Figure 2 shows the results of this experiment.

The left graph shows the run-time behavior over different training set sizes of C4.5 invoked with its default parameters versus C4.5 invoked with one pass of windowing (parameter setting `-t 1`). No significant differences can be observed, although windowing seems to eventually achieve a little run-time gain. The graph is quite similar to one resulting from experiments with ID3 and windowing shown by Wirth and Catlett (1988), so that we believe the differences in the original version of windowing and the one implemented in C4.5 are negligible in this domain. The results are also consistent with those of Quinlan (1993), who, for this domain, reported run-time savings of no more than 15% for windowing with appropriate parameter settings. In any case, it is obvious that the run-time of both C4.5 and C4.5 with windowing grows linearly with the number of training examples.[2]

On the other hand, the right half of Figure 2 depicts a similar experiment with windowing and DOS, a simple separate-and-conquer rule learning algorithm. DOS[3] is basically identical to pFoil, a propositional version of Foil (Quinlan, 1990) that uses information gain as a search heuristic, but no stopping criterion. Our implementation of windowing closely follows

2. Such a linear growth was already observed by Quinlan (1979) in some early experiments with windowing.
3. Dull Old Separate-and-conquer.

| rule | examples | perc. |
|---|---|---|
| `A = C, B = D.` | 4,096 | 1.563% |
| `C = E, D = F.` | 4,032 | 1.538% |
| `adjacent(A,E), adjacent(B,F).` | 30,072 | 11.472% |
| `C = E, A \= C.` | 22,932 | 8.748% |
| `D = F, B \= D.` | 22,932 | 8.748% |
| `C = E, not between(B,D,F).` | 1,456 | 0.555% |
| `D = F, not between(A,C,E).` | 1,456 | 0.555% |

Table 1: Coverage of a correct theory for the KRK domain.

the algorithm depicted in Figure 1. The results show a large difference between DOS without windowing, which again takes linear time, and DOS with windowing, which seems to take only constant time for training set sizes of about 3000 examples and above. This surprising constant-time learning behavior can be explained by the fact that if windowing is able to reliably discover a correct theory from a subset of the examples, it will never use more than this amount of examples for learning this theory. The remaining examples are only used for verifying that the learned theory has been correctly learned. These costs, albeit linear in principle, are negligible compared to the costs of learning the correct theory.

### 3.2 Windowing versus Random Sub-sampling

Of course, the intuitive arguments for the constant-time learning behavior of windowing could equally well apply to the decision tree learning method. However, the experimental evidence seems to show that in that case the learning time is at least linear in the number of examples. Why does windowing work better with rule learning algorithms than with decision tree learning algorithms?

To understand this behavior, let us take a closer look at the characteristics of windowing. Contrary to random sub-sampling, which will retain the distribution of examples of the original training set in the subsample, windowing purposefully tries to skew this example distribution by adding only examples that are misclassified by the current theory. As an example, consider the KRK illegality domain, where the task is to discriminate legal from illegal positions in a king-rook-king chess endgame, given features that compare the coordinates of two pieces and determine whether they are equal or adjacent, or whether one is between the others.

Table 1 shows the body of 7 rules that form a correct theory for illegal KRK positions in a first-order logic representation. The head of each rule is the literal `illegal(A,B,C,D,E,F)`. Next to the rules we give the number of different positions that are covered by the rule, along with their percentage out of the total of 262,144 different examples. About a third of the positions are illegal.[4] One can see that rules 3, 4, and 5 each cover more than 20,000 examples. They are usually easily found from small random samples of the dataset. On the other hand, rules 1 and 2, and in particular 6 and 7 cover a significantly smaller number

4. Note, however, that not all different positions form a different feature vector with the specified features as has been pointed out by Pfahringer (1995).

of examples. In order to obtain enough examples for learning these rules, one has to take a much larger random sample of the data.[5] This problem is closely related to the *small disjuncts problem* discussed by Holte, Acker, and Porter (1989).

How does windowing deal with this situation? Recall that it starts with a small random subsample of the available training data, and successively adds examples that are misclassified by the previously learned theory to this window. By doing so it *skews* the distribution of examples in a way that increases the proportion of examples covered by rules that are hard to learn, thereby decreasing the proportion of examples covered by rules that are easy to learn. Thus for each individual rule in the target theory, windowing tries to identify a minimum number of examples from which the rule can be learned.

### 3.3 Rule Learning versus Decision Tree Learning

Let us now return to the question of why windowing seems to work better with rule learning algorithms. We believe that its way of skewing the example distribution has different effects on divide-and-conquer decision tree learning algorithms and on separate-and-conquer rule learning algorithms. The nature of this difference lies in the way a condition is selected. Typically, a separate-and-conquer rule learner selects the test that maximizes the number of covered positive examples and at the same time minimizes the number of negative examples that pass the test. It usually does *not* pay any attention to the examples that do not pass the test. The selection of a condition in a divide-and-conquer decision tree learning algorithm, on the other hand, tries to optimize for *all* outcomes of the test. For example, the decision tree learner C4.5 (Quinlan, 1993) selects a condition by minimizing the *average* entropy over all its outcomes,[6] while the original version of the rule learner CN2 (Clark & Niblett, 1989) selects the test that minimizes the entropy for all examples that pass the selected test.

The consequence of this difference for the windowing process is that additional examples in the window of a decision tree learner will have a strong influence on the selection of tests at all levels of the tree. In particular, all examples have an equal influence on the selection of the root attribute. Thus, windowing's way of skewing the example distribution may cause significant changes in the learned tree. Separate-and-conquer rule learning algorithms, on the other hand, evaluate a rule using only the examples that are covered by it. As no examples will be added for an already correctly learned rule, the evaluation of this rule will not change if more examples are added to the window by incorrect rules. There might be some changes in the evaluation of some of its tests, because an incomplete rule can cover some of the uncovered positive examples or covered negative examples that were added in the last iteration of the windowing algorithm, but in general we think that this influence is not nearly as significant as for decision tree learning algorithms, where it is certain that each added example will influence the evaluation of the tests at the root of the tree.

5. This is even worse in the usual representation of this problem (Muggleton, Bain, Hayes-Michie, & Michie, 1989), where instead of the `between` relation, a `less_than` predicate is given in the background knowledge. In this representation, which will be used in all our experiments, rules 6 and 7 have to be reformulated using the `less_than` predicate, which is only possible if each of them is replaced with two new rules.
6. Actually, C4.5 maximizes information gain, which is computed by subtracting the average entropy from an information value that is constant for all considered tests.

More evidence for the sensitivity of ID3 to changes in the example distribution also comes from some experiments with C4.5 (Quinlan, 1993), where it turned out that changing windowing in a way that makes the class distribution in the initial window as uniform as possible produces better results.[7] As discussed above, we believe that this sensitivity is caused by the need of decision tree algorithms to optimize the class distribution in all successor nodes of an interior node. On the other hand, different class distributions will only have a minor effect on separate-and-conquer learners, because they are learning one rule at a time. Adding uncovered positive examples to the current window will not alter the evaluation of rules that do not cover the new examples, but it may cause the selection of a new root node in decision tree learning. We hypothesize from these deliberations that for windowing, rule learning algorithms exhibit more stability than decision tree learning algorithms. This, in turn, can lead to better run-times because the newly added examples will not interact with the parts of theory that have already been learned well.

### 3.4 Domain Redundancy

It is clear that windowing requires a *redundant* training set in order to work effectively. Intuitively, we say a training set is redundant, if it contains more examples that are actually needed for inducing the correct domain theory. If this is not the case, i.e., if every example is needed for inferring a correct theory, windowing is unlikely to work well, because eventually all examples have to be added to the window and there is nothing to be gained.

A computable measure for the redundancy of a given domain would enable us to evaluate the potential effectiveness of windowing in this domain. Unfortunately, we do not know of many approaches that deal with this problem. A notable exception is the work by Møller (1993) where the use of the *conditional population entropy (CPE)* is suggested for measuring the redundancy of a domain. The CPE is defined as

$$CPE = -\sum_{i=1}^{n_c} p(c_i) \sum_{a=1}^{n_a} \sum_{v=1}^{n_{v_a}} p(x_{a,v}|c_i) \log p(x_{a,v}|c_i) \tag{1}$$

where $n_c$ is the number of classes, $n_a$ is the number of attributes and $n_{v_a}$ is the number of values for attribute $a$. $c_i$ stands for the $i$-th class and $x_{a,v}$ represents the $v$-th value of attribute $a$. One can interpret the formula as computing the sum of the respective average entropies of the class variable in one-level decision trees for predicting each attribute. Møller normalizes the CPE as follows:

$$Red = 1 - \frac{CPE}{\sum_{a=1}^{n_a} \log n_{v_a}} \tag{2}$$

The normalization factor is the maximum value that the CPE can assume, which is when $p(x_{a,v}|c_i) = \frac{1}{n_{v_a}}$ and $p(c_i) = \frac{1}{n_c}$. Using this normalization, redundancy can be measured with a value between 0 and 1, 1 meaning high redundancy, 0 meaning no redundancy.

However, it is unclear how this measure corresponds to our intuitive notion of redundancy. For example, two example sets of a domain that differ only in the fact that the

7. This equal-frequency sub-sampling method has first been described by Breiman, Friedman, Olshen, and Stone (1984) for dynamically drawing a subsample at each node in a decision tree from which the best split for this node is determined. In C4.5 it is used to seed the initial window.

second contains each example of the first set twice, would have exactly the same redundancy estimate. Intuitively, however, the second example set should be more redundant than the first.

A radically different approach to defining redundancy (in a somewhat different context) was undertaken by Gamberger and Lavrač (1997): They call a training set *n-saturated*, iff there is no subset of size $n$ whose removal would cause a different theory to be learned. They propose to use the notion of $n$-saturation for approximating the concept of *saturation*, which intuitively means that the training set contains enough examples for inducing the correct target hypothesis (Gamberger & Lavrač, 1997). This notion corresponds quite closely to our intuitive concept of redundancy. However, the computational complexity of testing for $n$-saturation can be quite high.

More importantly, the notion of saturation as defined above does not take into account that different regions of the example space may have different degrees of redundancy. We have already seen in Table 1 that often the rules in a target theory have different degrees of coverage. Consequently, a randomly chosen training set will typically contain more examples that are covered by a high-coverage rule of the target theory than examples that are covered by a low-coverage rule. In other words, the subset of the examples from which the high-coverage rules are learned can already be redundant, while the subset from which the low-coverage rules should be learned does not yet contain enough examples for inducing the correct rules. In Gamberger and Lavrač's notion of saturation, such a training set would be considered non-saturated.

Windowing tries to exploit these different degrees of redundancy in a training set. If some parts of the example space are already covered by correct rules, no more examples of these regions will be added to the window. We have already discussed that this skewing of the example distribution is more appropriate for separate-and-conquer rule learning algorithms than for divide-and-conquer decision tree learning algorithms because the former learn each rule independently. In the next section, we will discuss a new windowing algorithm for rule learning systems that aims at further exploiting this property.

## 4. Integrative Windowing

One thing that happens frequently when using windowing with a rule learning algorithm is that good rules have to be discovered again and again in subsequent iterations of the windowing procedure. Although correctly learned rules will add no more examples to the current window, they have to be re-learned in the next iteration as long as the current theory is not complete and consistent with the entire training set. We have developed a new version of windowing, which tries to exploit the fact that regions of the example space that are already covered by good rules need not be further considered in subsequent iterations. Because of its technique of successively integrating learned rules into the final theory, we have named our method *Integrative Windowing*.

### 4.1 The Algorithm

The algorithm shown in Figure 3 starts just like basic windowing: it selects a random subset of the examples, learns a theory from these examples, and tests it on the remaining examples. However, contrary to basic windowing, it does not merely add incorrectly classified examples

```
procedure IntegrativeWindowing(Examples,InitSize,MaxIncSize)

Window = RandomSample(Examples,InitSize)
Test = Examples \ Window
OldRules = ∅
repeat
    NewRules = Induce(Window)
    Theory = NewRules ∪ OldRules
    NewWindow = ∅
    OldTest = ∅
    for Example ∈ Test
        Test = Test \ Example
        if Classify(Theory,Example) ≠ Class(Example)
            NewWindow = NewWindow ∪ Example
        else
            OldTest = OldTest ∪ Example
        if |NewWindow| = MaxIncSize
            exit for
    Test = Append(Test,OldTest)
    Window = Window ∪ NewWindow ∪ Cover(OldRules)
    OldRules = ∅
    for Rule ∈ Theory
        if Consistent(Rule,NewWindow)
            OldRules = OldRules ∪ Rule
            Window = Window \ Cover(Rule)
until NewWindow = ∅
return(Theory)
```

Figure 3: Integrative Windowing.

to the window for the next iteration, but also removes examples from the window if they are covered by consistent rules. A rule is considered consistent, when it did not cover a negative example during the testing phase. Note that this does not necessarily mean that the rule is consistent with all examples in the training set because it may contradict an example that has not yet been tested at the point where *MaxIncSize* misclassified examples have been found. Thus apparently consistent rules have to be remembered and tested again in the next iteration. However, testing is much cheaper than learning, so we expect that removing the examples that are covered by these rules from the window should keep the window size small and thus decrease learning time.

## 4.2 Implementation

To test this hypothesis, we implemented several algorithms in Common LISP, building on Ray Mooney's publicly available Machine Learning library. The implemented algorithms are DOS, a basic separate-and-conquer rule learning algorithm using information gain as a search heuristic and no stopping criterion for noise handling, WIN-DOS-3.1, an implementation of windowing as shown in Figure 1 that wraps a windowing procedure around DOS, and WIN-DOS-95, an algorithm that integrates windowing into DOS as shown in Figure 3. All algorithms are limited to 2-class problems, i.e., they learn a theory that discriminates positive from negative examples by classifying all examples that are covered by the rules as positive, and all examples that are not covered by the rules as negative. Note, however, that this is not a principle limitation of the approach, as there are several ways for solving multi-class problems with binary rule learning algorithms of the type discussed in this paper (Clark & Boswell, 1991; Ali & Pazzani, 1993; Dietterich & Bakiri, 1995).

In preliminary experiments it turned out that one problem that happens more frequently in integrative windowing than in regular windowing or basic separate-and-conquer learning is that of over-specialized rules. Often a consistent rule is found at a low example size, but other rules are found later that cover all of the examples this special rule covers. Note that this problem cannot be removed with a syntactic generality test: consider, for example, the case where a rule stating that a KRK position is illegal if the two kings are on the same square is learned from a small set of the data, and a more general rule is discovered later, which states that all positions are illegal in which the two kings occupy adjacent squares. Sometimes the examples of the special case can also be covered by more than one of the other rules.

We have implemented a heuristic procedure for removing such redundant rules: After the final theory has been learned by either of the algorithms, each of its rules is tested on the complete training set and the rules are ordered according to the number of examples they cover. Starting with the rule with the least coverage, each rule is tested whether the examples it covers are also covered by the remaining rules. If so, the rule is removed. This procedure can be implemented quite efficiently and will only be performed once at the end of each of the three algorithms.

### 4.3 Experimental Setup

We compared both versions of windowing on a variety of noise-free domains. In each domain we ran a series of experiments with varying training set sizes. For each training set size, 10 different subsets of this size were selected from the entire set of preclassified examples. All three algorithms, DOS, Win-DOS-3.1, and Win-DOS-95 were run on each of these subsets and the results of the 10 experiments were averaged. For each experiment we measured the accuracy of the learned theory on the entire example set and the total run-time of the algorithm.[8] To have a more reliable complexity measure than the implementation-dependent run-time, we also measured the total number of examples that were processed by the basic learning algorithm. For DOS, this number is identical to the size of the respective training set, while for the windowing algorithms it is computed as the sum of the training set sizes of all iterations of windowing. For the noise-free case, it turned out that this factor determined the run-time of the algorithms, so that its graphs were almost identical to the graphs for run-time results. Therefore, for reasons of space efficiency, we will only present the run-time curves (except in Figure 4).[9]

All experiments shown below were conducted with a setting of *InitSize* = 100 and *MaxIncSize* = 50. These settings are briefly discussed in section 4.6.

8. Measured in CPU seconds of a microSPARC 110MHz running compiled Allegro Common Lisp code under SUN Unix 4.1.3.
9. We did not compute this measurement for the experiments in noisy domains (section 5) because the way we compute the completeness check (section 5.2.2) invalidates these measurements.

| Domain | Size | `c4.5 -t1` vs. `c4.5` | Redundancy |
|---|---|---|---|
| Mushroom | 8,124 | 98.8 % | 46.61 % |
| KRKN | 10,000 | 91.2 % | 46.05 % |
| KRKP | 3,196 | 112.8 % | 43.81 % |
| KRK (prop.) | 10,000 | 113.8 % | 21.88 % |
| Tic-Tac-Toe | 958 | 258.0 % | 4.15 % |
| Binary Shuttle | 43,500 | 55.3 % | — |

Table 2: Domains used in the experiments, along with their size, the performance of C4.5 with windowing versus C4.5 without windowing, and an estimate for the redundancy of the domain.

### 4.4 Domains

We evaluated the algorithms on a variety of reasonably large and noise-free training sets from the UCI collection of Machine Learning databases. As our implementation can only handle 2-class problems, we constructed a binary version of the multi-class *Shuttle* domain by discriminating examples of majority class from all other classes. In the KRK illegality domain we used a propositional version of the original relational learning problem (Muggleton et al., 1989), where each position is encoded with features that correspond to the truth values of the 18 different meaningful instantiations of the `adjacent`, `equal`, and `less_than` relations in the background knowledge.

Table 2 shows the total number of examples available for each domain and the ratio of the average run-time of C4.5 with windowing (invoked using the parameter setting `-t 1`) versus C4.5 without windowing. The last column shows the redundancy of the domain, estimated with Møller's conditional population entropy heuristic (2). Interestingly enough, there seems to be a (negative) correlation between the performance of C4.5's windowing algorithm and this redundancy measure.[10] In general, the results with C4.5 confirm the results of Wirth and Catlett (1988) that not much can be gained with the use of windowing for ID3-like learners. The only exception is the *Shuttle* domain, where windowing can save almost half of C4.5's run-time.

### 4.5 Results

Figure 4 shows the accuracy, the number of processed examples, and the run-time results for the three algorithms in the *Mushroom* domain. Win-DOS-3.1 seems to be effective in this domain, at least for larger training sets (> 1000). Our improved version, Win-DOS-95, clearly outperforms simple windowing in terms of run-time, while there are no significant differences in terms of accuracy.

In a typical run with the above-mentioned parameter settings, Win-DOS-3.1 needs about 3 to 5 iterations for learning the correct concept, the last of them using a window size of about 200 to 350 examples. Win-DOS-95 needs about the same number of iterations,

10. The measure could not be computed for the *Shuttle* domain, because this domain contains numerical attributes, which cannot be handled in a straight-forward fashion.

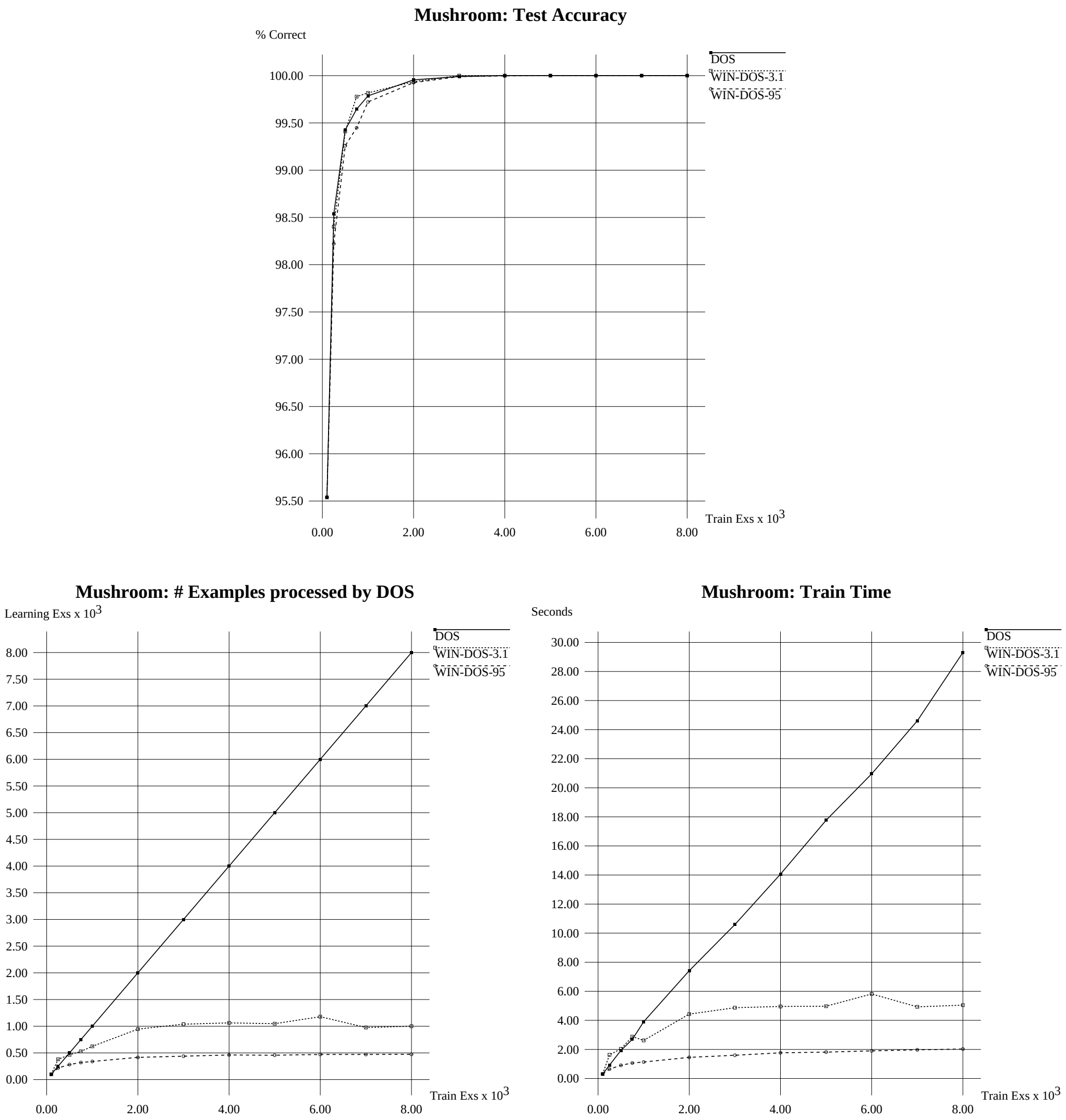

Figure 4: Results in the *Mushroom* domain.

but it rarely keeps more than 150 examples in its window. In total, WIN-DOS-3.1 submits about 1000 examples to the DOS learner, while WIN-DOS-95 can save about half of them. Figure 4 shows that these numbers remain almost constant after a saturation point of about 3,000 examples is reached, which is the point where all learners learn 100% correct theories. Learning from 500 to 1000 randomly selected examples does not result in correct theories, as can be inferred from the DOS accuracy curve. Thus, in this domain, a performance gain comparable to that achieved by windowing cannot be achieved with random sub-sampling. Hence we conclude that the systematic sub-sampling performed by windowing has its merits.

Figure 5 shows the accuracy and run-time results for the three algorithms in the *KRK* and *KRKN* domains. The picture here is quite similar to Figure 4: WIN-DOS-3.1 seems to be effective in both domains, at least for larger training set sizes. Our improved version, WIN-DOS-95, clearly outperforms basic windowing in terms of run-time, while there are no significant differences in terms of accuracy. In the KRK domain, predictive accuracy reaches 100% at about 5000 training examples for all three algorithms. At lower training set sizes, WIN-DOS-95 is a little ahead, but in general there are no significant differences. The run-time of both windowing algorithms reaches a plateau at about the same size of 5000 training examples, which again shows that windowing does not use additional examples once it is able to discover a correct theory from a certain sample size.

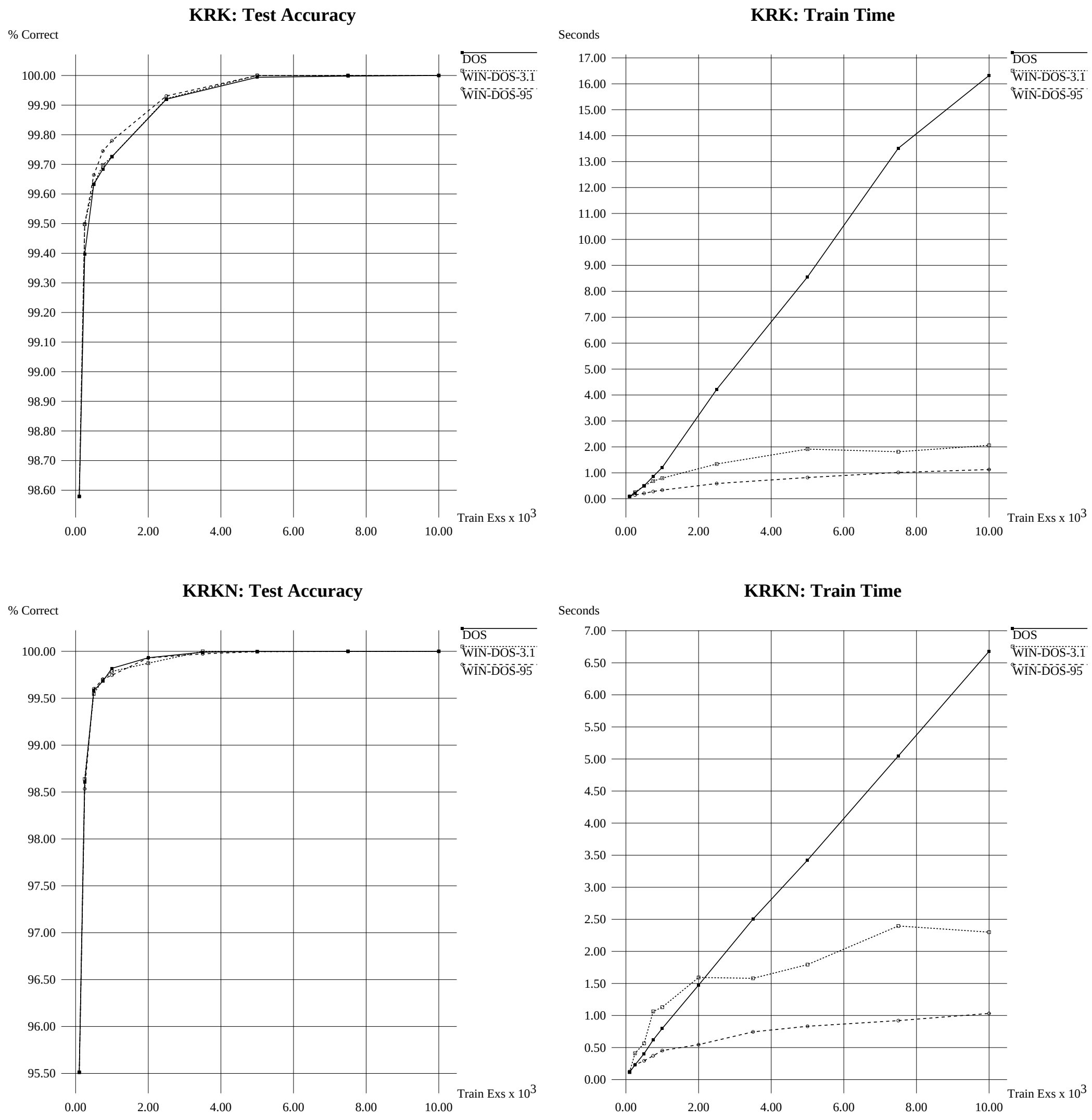

Figure 5: Results in the KRK, and KRKN domains.

The results in the KRKN chess endgame domain with training set sizes of up to 10,000 examples are quite similar. However, for smaller training set sizes, which presumably do not contain enough redundancy, basic windowing can take significantly longer than learning from the complete data set.

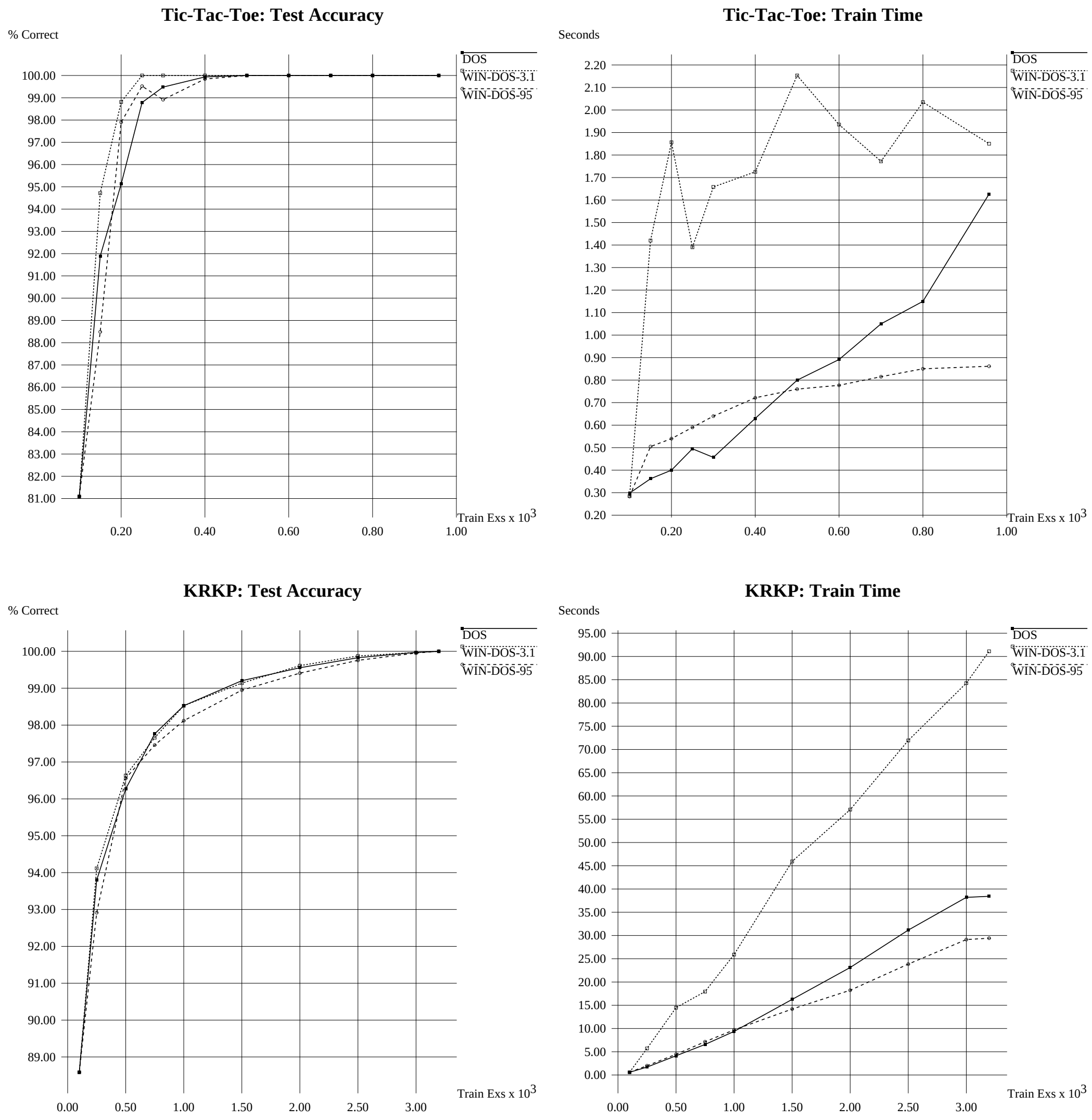

Figure 6: Results in the Tic-Tac-Toe and KRKP domains.

This behavior is more obvious in the results in the *Tic-Tac-Toe* endgame domain[11] shown in the top portion of Figure 6. Here, the predictive accuracy of all three algorithms reaches 100% at about half the total example set size. Interestingly, WIN-DOS-3.1 reaches this point considerably earlier (at about 250 examples). On the other hand, WIN-DOS-3.1 is not able to achieve an advantage over DOS in terms of run-time, although it seems quite likely that it would overtake DOS at slightly larger training set sizes. WIN-DOS-95 reaches this break-even point much earlier and continues to build up a significant gain in run-time at larger training set sizes.

11. Note that this example set is only a subset of the Tic-Tac-Toe data set that has been studied by Wirth and Catlett (1988). We did not have access to the full data set.

In all domains considered so far, removing a few randomly chosen examples from the larger training sets did not affect the learned theories. Intuitively, we would call such training sets *redundant* as discussed in section 3.4. In the 3196 example KRKP data set, on the other hand, the algorithms are not able to learn theories that are 100% correct when tested on the complete data set unless they use the entire data set for training. Note that in this case the 100% accuracy estimate is in fact a re-substitution estimate, which may be a bad approximation for the true accuracy. We would call such a data set *non-redundant*, as it seems to be the case that randomly removing only a few examples will already affect the learned theories. In this domain, Win-DOS-3.1 processes about twice as many examples as DOS for each training set size. Our improved version of windowing, on the other hand, processes only a few more examples than DOS at lower sizes, but seems to be able to exploit some redundancies of the domain at larger training set sizes. We interpret this a evidence for our hypothesis that even in non-redundant training sets, some parts of the example space may be redundant (section 3.4). This is consistent with previous findings, which showed that in this domain, a few accurate rules with high coverage can be found easily from small training sets, while the majority of the rules have low coverage and are very error-prone (Holte et al., 1989). Interestingly enough, however, Møller's redundancy estimate, which seemed to correlate well with C4.5's performance in these domains in Table 2, is a poor predictor for the performance of our windowing algorithms. The KRKP set, which exhibits the worst performance and is not redundant according to our definition, has a much better redundancy estimate than the KRK data set, for which windowing works much better.

Figure 7 shows the results in the 43,500 examples *Shuttle* domain. For this domain, a separate test set is available (14,500 examples). The accuracy graphs show the performance of the algorithms on this test set. The upper half shows the results when training on the minority class (i.e., learning a theory for all examples with a non-1 classification). Both windowing algorithms perform exceptionally well in this domain, with Win-DOS-95 being about twice as fast as Win-DOS-3.1. In terms of accuracy, Win-DOS-3.1 is more accurate than both other algorithms, Win-DOS-95 being a little ahead of DOS. As this example set has a relatively skewed class distribution (about 80% of the examples are of class 1), we decided to also try the algorithms on the reverse problem by swapping the labels of the positive and negative examples. The results in terms of run-time did not change. However, in terms of accuracy, the performance of DOS improved, so that it equals Win-DOS-3.1 with Win-DOS-95 being behind both. We have not yet found a good explanation for this phenomenon.

It is interesting to note that this is the only domain we have tried that contains numeric attributes. We handled these in the common way using threshold tests. The candidate thresholds for each test are selected between changes in the class variables in the sorted list of values of the continuous attributes (Fayyad & Irani, 1992).[12] As the windowing algorithms will typically have to choose a threshold from a much smaller example set, they are likely to choose different thresholds. Whether this has a positive (less over-fitting), negative (the optimal threshold may be missed), or no effect at all is an open question, which has already been raised by Breiman et al. (1984) and has been further explored in Catlett's work on *peepholing* (Catlett, 1992).

12. The sorting causes the slightly super-linear slope in the run-time curves of DOS.

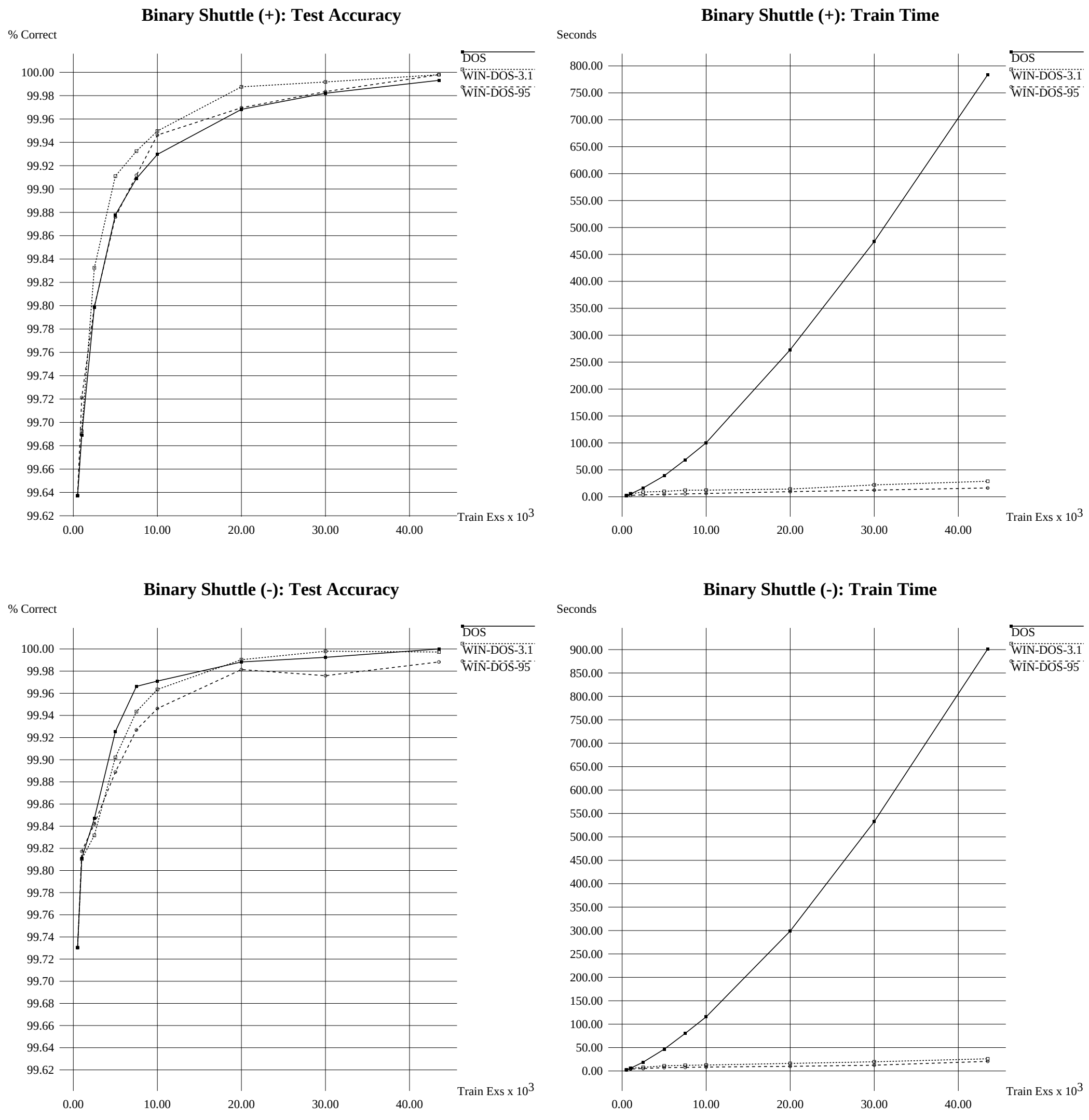

Figure 7: Results in the binarized *Shuttle* domain.

In summary, the results have shown us that significant run-time gains can be achieved with windowing (usually without losing accuracy), and even more so with integrative windowing in a variety of noise-free domains. However, we have also seen that redundancy is in fact a crucial condition for the effectiveness of windowing, so that a further exploration of domain redundancy is suggested as a promising area for further research.

### 4.6 A Note on Parameter Settings

All experiments reported above were performed with *InitSize* = 100 and *MaxIncSize* = 50. Different variations of the *InitSize* parameter have been investigated by Wirth and Catlett (1988). Their results indicate that the algorithm is quite insensitive to this parameter, which we have empirically confirmed. Nevertheless, it might be worthwhile to use a theoretical analysis of sub-sampling similar to the one performed by Kivinen and Mannila (1994) for determining promising values of this parameter for a particular domain.

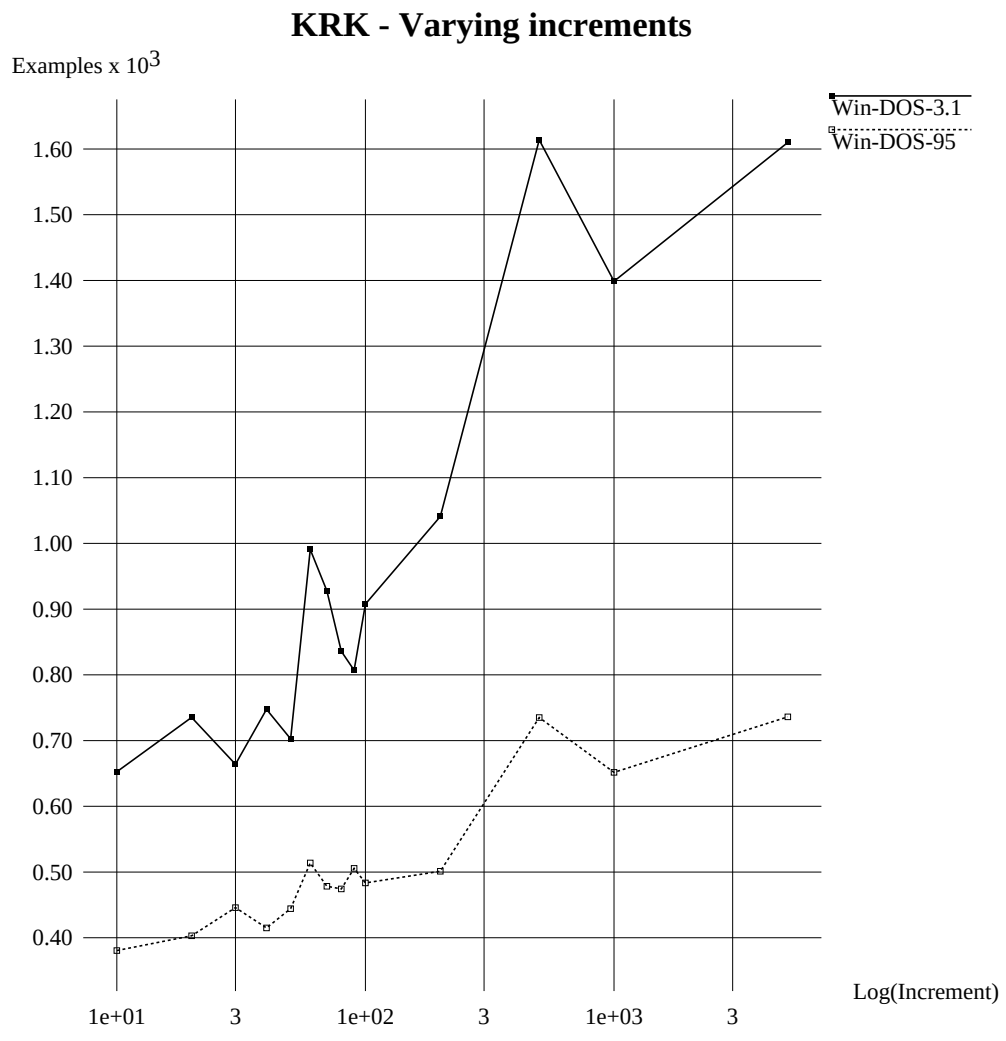

Figure 8: Number of examples processed by the windowing algorithms for varying values of their *MaxIncSize* parameter (shown on a logarithmic scale).

However, we attribute more significance to the choice of the *MaxIncSize* parameter, which specifies the maximum number of examples that can be added to a window. Figure 8 shows the results of experiments with the 10,000 examples set of the KRK domain, in which we varied *MaxIncSize* from 10 to 5000 (plotted on a logarithmic scale on the $x$-axis). The performance of the algorithms in terms of number of processed examples is best if this parameter is kept comparably low. In the range of 10 to 50 examples, the parameter is relatively insensitive to its exact setting. If more examples are added to the window, performance degrades. For example at *MaxIncSize* = 50, Win-DOS-3.1 performs about 4 iterations of the basic learning algorithm processing a total of about 700 examples, the final window containing about 250 examples. At *MaxIncSize* = 1000 on the other hand, the basic learning module not only has to process about twice as many examples, but windowing also takes more iterations to converge. Similar behavior can be observed for Win-DOS-95. Thus it seems to be important to continuously evaluate the learned theories in order to focus the learner on the parts of the search space that have not yet been correctly learned. This finding contradicts the heuristic that is currently employed in C4.5, namely to add at least half of the total misclassified examples. However, this heuristic was formed in order to make windowing more effective in noisy domains (Quinlan, 1993), a goal that in our opinion cannot be achieved with merely using a noise-tolerant learner inside the windowing loop, for reasons discussed in the next section.

## 5. Windowing and Noise

So far we have only dealt with noise-free domains. While we believe that there are many rewarding learning tasks that can be successfully attacked with noise-free learning algorithms,[13] noise is commonly considered as a typical property of real-world data. In this section we will have a closer look at the behavior of windowing with noisy data, analyze the problems, and discuss a possible solution.

### 5.1 Noise is a Problem

An efficient adaptation of the basic windowing technique shown in Figure 1 to noisy domains is a non-trivial endeavor. In particular, it cannot be expected that the use of a noise-tolerant learning algorithm inside the windowing loop will lead to performance gains in noisy domains. In our opinion, the main problem with windowing in a noisy domain lies in the fact that a good theory will misclassify most of the noisy examples, and consequently incorporate them into the learning window for the next iteration. On the other hand, the window will typically only contain a subset of the original learning examples. Hence, after a few iterations, the proportion of noisy examples in the learning window can be much higher than the noise level in the entire data set. Naturally, this makes the task for the learning module considerably more difficult.

Assume, for example, that your favorite noise-tolerant learner has learned a correct theory from a randomly selected starting window of size 1000 in a 11,000 examples domain. Further assume that 10% of the examples are labeled incorrectly. Therefore, the correct theory will misclassify 1000 of the remaining 10,000 examples because they are noisy. These examples will consequently be added to the window, thus doubling its size. Assuming that the original window also contained about 10% noise, more than half of the examples in the new window are now erroneous, so that the classification of the examples in the window is in fact random. It can be assumed that many more examples have to be added to the window in order to recover the structure that is inherent in the data. This conjecture is consistent with the experimental results of Wirth and Catlett (1988) and Catlett (1991a), which showed that windowing is highly sensitive to noise.

### 5.2 Towards Noise-Tolerant Windowing

The integrative windowing algorithm described in section 4, which is only applicable to noise-free domains, is based on the observation that rule learning algorithms will re-discover good rules again and again in subsequent iterations of the windowing procedure. Such consistent rules do not add examples to the current window (hence they are unlikely to change), but they nevertheless have to be re-discovered in subsequent iterations. Integrative windowing detects these rules early on, saves them, and removes all examples they cover from the window, thus gaining computational efficiency.

In order to adapt this procedure for noisy domains, three parts of the algorithm have to be modified:

13. Think, e.g., of Ken Thompson's 3 CD-Roms of chess endgames, to which every competitive chess program has an interface. A compression of these endgames into a simple set of rules would certainly be appreciated by the computer chess industry (Fürnkranz, 1997b).

**Consistency-Check:** *When is a rule learned from the window good enough?*

In the noise-free case, all rules that do not cover any negative examples are added to the final theory. For noisy data, a criterion has to be found that allows rules to cover some noisy negative examples.

**Completeness-Check:** *When should we stop adding rules to a theory?*

In the noise-free case, rules are added to the theory until all positive examples are covered by at least one rule. For noisy data, we have to find a criterion that estimates whether the remaining uncovered positive examples can be considered as noise or whether another rule should be added to the theory that explains (some of) them.

**Resampling:** *Which examples should be added to the current window?*

In the noise-free case, all misclassified examples are candidates for being added to the window. However, we have seen above that this can lead to severe problems with windowing, because noisy examples will be misclassified by a correct theory.

We have built a prototype system that deals with these three questions in the way described in the following sections.

#### 5.2.1 Consistency-Check

The first choice we have to make is to find a criterion for determining which rules should be added to the final theory, and which rules should be further improved. A straight-forward idea is to use some sort of significance test in order to determine whether a rule that has been learned from the current window is significant on the entire set of training examples. We have experimented with a variety of criteria known from the literature, but found that they are insufficient for our purposes. For example, it turned out that, at higher training set sizes, CN2's likelihood ratio significance test (Clark & Niblett, 1989), which accepts a rule when the distribution of positive and negative examples covered by the rule differs significantly from the class distribution in the entire set, will deem any rule learned from the window as significant.

Eventually, we have settled for the following criterion: For each rule $r$ learned from the current window we compute two accuracy estimates, $AccWin(r)$ which is determined using only examples from the current window and $AccTot(r)$ which is estimated on the entire training set. The criterion we use for detecting good rules consists of two parts:

1. The $AccWin(r)$ estimate has to be significantly above the default accuracy of the domain. This is ensured by requiring that

$$AccWin(r) - SE(AccWin(r)) > DA \tag{3}$$

where $DA$ is the default accuracy, and $SE(p) = \sqrt{\frac{p(1-p)}{n}}$ is the standard error of classification (Breiman et al., 1984).

2. The accuracy of the learned rule on the window has to be within a certain interval of its overall accuracy:[14]

$$|AccWin(r) - AccTot(r)| \leq \alpha \times SE(AccTot(r)) \quad (4)$$

$\alpha \geq 0$ is a user-settable parameter that can be used to adjust the width of this range.

The purpose of the first heuristic is to avoid rules with a bad classification performance (in particular it weeds out many rules that have been derived from too few training examples), while the second criterion aims at making sure that the accuracy estimates that were computed on the current window (and were used in the heuristics of the learning algorithm) are not too optimistic compared to the true accuracy of the rule, which is approximated by the accuracy measured on the entire training set.

The parameter $\alpha$ determines the degree to which the estimates $AccWin(r)$ and $AccTot(r)$ have to correspond. A setting of $\alpha = 0$ requires that $AccWin(r) = AccTot(r)$, which in general will only be happen if $r$ is consistent or has been learned from the entire training set. This is the recommended setting in noise-free domains. In noisy domains, values of $\alpha > 0$ have to be used because the rules returned from the learning algorithm will typically be inconsistent on the training set. Note, however, that a setting of $\alpha = 0$ in a noisy domain will *not* lead to over-fitting and a decrease in predictive accuracy because over-fitting is caused by too optimistic estimates of a rule's accuracy. The chance of accepting such rules decreases with the value of $\alpha$. Clearly, if the chance of a rule being accepted decreases, the run-time of algorithm increases because windowing has to perform more iterations. The extreme case, $\alpha = 0$, causes the window to grow until it contains all examples. Then the rule is accepted by the second criterion because the examples used for estimating $AccWin(r)$ and $AccTot(r)$ are basically the same. Typical settings in noisy domains are $\alpha = 0.5$ or $\alpha = 1.0$. $\alpha = \infty$ will move all rules that have survived the first criterion into the final rule set. In as much as the learner is sufficiently noise-tolerant and the initial example size is sufficiently large, the algorithm implements learning from a random subsample if $\alpha = \infty$.

### 5.2.2 Completeness-Check

A straight-forward approach for attacking the completeness problem, i.e., the decision when to stop learning more rules, would be to stop learning whenever the learner can find no more rules from the current window. This approach, however, might miss some important rules that could be found if only more of the uncovered positive examples had been added to the window. So a different approach is to continue the windowing process until all remaining positive examples are in the learning window. This, on the other hand, may lead to many unnecessary iterations in the case when no more meaningful rules can be found.

We aimed at a compromise here. The strategy we employ is to double the window size in the case when the learner has not discovered any rules from the current window, which ensures a fast convergence for the case when no more rules can be found, but also tries to find the rules from lower window sizes first. Also note that this approach relies on a truly

14. Note that (4) differs from the version of (Fürnkranz, 1997d) by using $SE(AccTot(r))$ instead of $SE(AccWin(r))$ on the right hand side. We have found this version to work somewhat more reliably.

noise-tolerant learning algorithm. If, for example, the learning algorithm discovers rules even in randomly classified training data (i.e., pure noise), this criterion will never kick in. The windowing algorithm will then discover that these rules are insignificant and continue to add more examples to the window. This will go on until all examples are in the window, in which case the found rules will be retained.

#### 5.2.3 RESAMPLING

Our main concern with the resampling problem was that adding only misclassified examples is likely to increase the noise level inside the window. To avoid this, we form a set of candidates containing *all* examples that are not yet in the window and that are covered by insignificant rules, plus all uncovered positive examples. The algorithm then selects *MaxIncSize* of these candidate examples and adds them to the window. We stick to adding uncovered *positive* examples only, because after more and more rules have been discovered, the proportion of positive examples in the remaining training set will considerably decrease, so that the chances of randomly picking a positive example from the set of all uncovered examples would decrease, which in turn might slow down the learner. Although adding only positive uncovered examples may increase the chances of learning over-general rules, these will be discovered by the second part of our criterion and appropriate counter-examples will eventually be added to the window.

### 5.3 The Algorithm

Figure 9 shows the final algorithm. At the beginning it proceeds just like the basic or integrative windowing algorithms described earlier in this paper. It selects a random subset of the examples, learns a theory from these examples, and tests it on the remaining examples. Like integrative windowing, it does not merely add examples that have been incorrectly classified to the window for the next iteration, but it also removes all examples from this window that are covered by good rules. To determine good rules, it tests the individual rules that have been learned from the current window on the entire data set and performs the consistency check described in section 5.2.1 (procedure SIGNIFICANT). If the rule passes the test, it is added to the final theory, and all examples that are covered by it are removed from the training set (and the window). Otherwise, the examples covered by this rule become candidates for being added to the window in the next iteration. After all uncovered positive examples have also been added to this candidate set, the algorithm randomly selects *MaxIncSize* examples that are added to the window. Not shown in the algorithm is the completeness check described in section 5.2.2, which doubles the window size if the noise-tolerant learner does not find any rules.

With a setting of $\alpha = 0$, NOISETOLERANTWINDOWING is very similar to the INTEGRATIVEWINDOWING algorithm of Figure 3, with the difference that the latter only tests a theory until it has collected *MaxIncSize* new examples to add to the current window. Thus it cannot determine whether a rule has already been tested on all examples and has to test the stored rules in all subsequent iterations. NOISETOLERANTWINDOWING, on the other hand, tests a rule on the entire training set. If it finds the rule to be significant it will add it to the final rule set and will never test it again. Consequently, the examples covered by such a rule can be removed not only from the window, but from the entire training set.

```
procedure NoiseTolerantWindowing(Examples,InitSize,MaxIncSize)

Window = RandomSample(Examples,InitSize)
Theory = ∅
repeat
    NewRules = NoiseTolerantLearner(Window)
    NewWin = Window
    NewExs = Examples
    Candidates = ∅
    for Rule ∈ NewRules
        if Significant(Rule, Window,Examples)
            Theory = Theory ∪ Rule
            NewWin = NewWin \ Cover(Rule, Window)
            NewExs = NewExs \ Cover(Rule,Examples)
        else
            Candidates = Candidates ∪ Cover(Rule,Examples)
    for Example ∈ Positive(Examples)
        if Example ∉ Cover (NewRules,Examples)
            Candidates = Candidates ∪ Example
    Window = NewWin ∪ RandomSample(Candidates,MaxIncSize)
    Examples = NewExs
until Candidates = ∅
return (Theory)
```

Figure 9: A noise-tolerant version of integrative windowing.

### 5.4 Experimental Evaluation

We implemented the algorithm described in the last section in the same LISP environment as the noise-free algorithms. The noise-tolerant learning algorithm used was I-RIP, a separate-and-conquer learner halfway between I-REP (Fürnkranz & Widmer, 1994; Fürnkranz, 1997e) and RIPPER (Cohen, 1995). Like I-REP, it learns single rules by greedily adding one condition at a time (using Foil's information gain heuristic (Quinlan, 1990)) until the rule no longer makes incorrect predictions on the growing set, a randomly chosen set of 2/3 of the training examples. Thereafter, the learned rule is simplified by greedily deleting conditions as long as the performance of the rule does not decrease on the remaining set of examples (the pruning set). All examples covered by the resulting rule are then removed from the training set and a new rule is learned in the same way until all positive examples are covered by at least one rule or the stopping criterion fires. Like RIPPER, I-RIP evaluates the quality of a rule on the pruning set by computing the heuristic $\frac{p-n}{p+n}$ (where $p$ and $n$ are the covered positive and negative examples respectively) and stops adding rules to the theory whenever the fraction of positive examples that are covered by the best rule does not exceed 0.5. These choices have been shown to outperform I-REP's original choices (Cohen, 1995). I-RIP is quite similar to I-REP*, which is also described by Cohen (1995), but it retains I-REP's method of considering all conditions for pruning (instead of considering a final sequence of conditions).

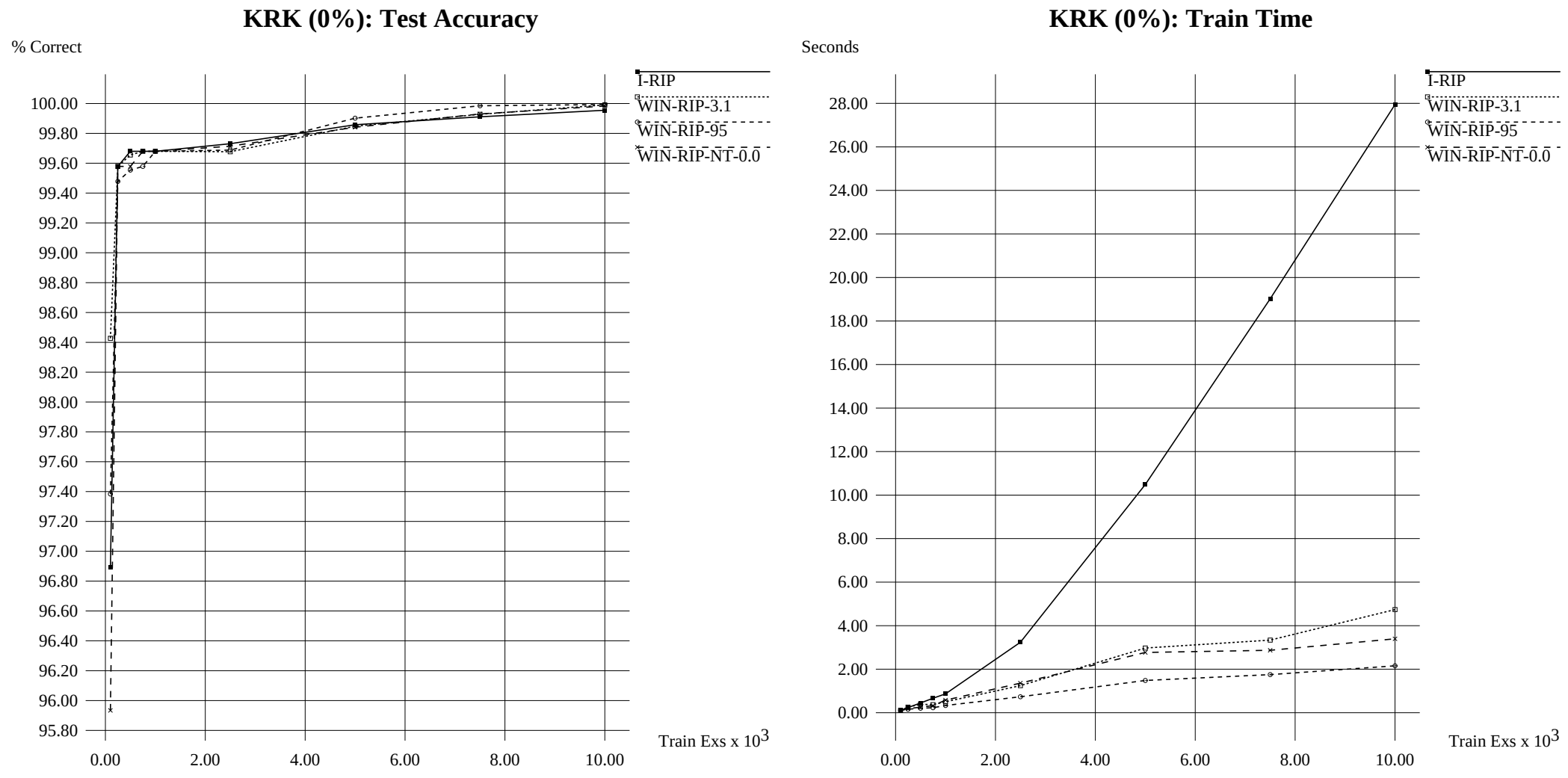

Figure 10: Results for the noise-free KRK domain.

Around I-RIP, we wrapped the windowing procedure of Figure 9. The resulting algorithm is referred to as Win-RIP-NT.[15] Again, the implementations of both algorithms remove semantically redundant rules in a post-processing phase as described in section 4.2.

We studied the behavior of the algorithms in the KRK illegality domain with varying levels of artificial noise. A noise level of $n\%$ means that in about $n\%$ of the examples the class label has been replaced with a random class label. In each of the experiments described in this section, we report the average results on 10 different subsets in terms of run-time of the algorithm and measured accuracy on a separate noise-free test set. We did not evaluate the algorithms according to the total number of examples processed by the basic learning algorithm, because the way we compute the completeness check (doubling the size of the window if no rules can be learned from the current window, see section 5.2.2) makes these results less useful. All algorithms were run on identical data sets, but some random variation results from the fact that I-RIP uses internal random splits of the training data. All experiments shown below were conducted with the same parameter setting of *InitSize* = 100 and *MaxIncSize* = 50 which performed well in the noise-free case. We have not yet made an attempt to evaluate their appropriateness for noisy domains.

First, we aimed at making sure that the noise-free setting ($\alpha = 0.0$) of the Win-RIP-NT algorithm still performs reasonably well, so that the noise-tolerant generalization of the IntegrativeWindowing algorithm does not lose its efficiency in noise-free domains. Figure 10 shows the performance of I-RIP, Win-RIP-3.1 (basic windowing using I-RIP in its loop), Win-RIP-95 (integrative windowing using I-RIP), and Win-RIP-NT-0.0 (the noise-tolerant version of integrative windowing with a setting of $\alpha = 0.0$). The graphs resemble very much the graphs of Figure 5 for the KRK domain, except that I-RIP over-generalizes occasionally, so that the accuracy never reaches exactly 100%. Win-RIP-NT-0.0 performs a little worse than Win-RIP-95, but it still is better than regular windowing. A similar

15. NT, of course, stands for Noise-Tolerant. Special thanks to the reviewer who suggested this notation.

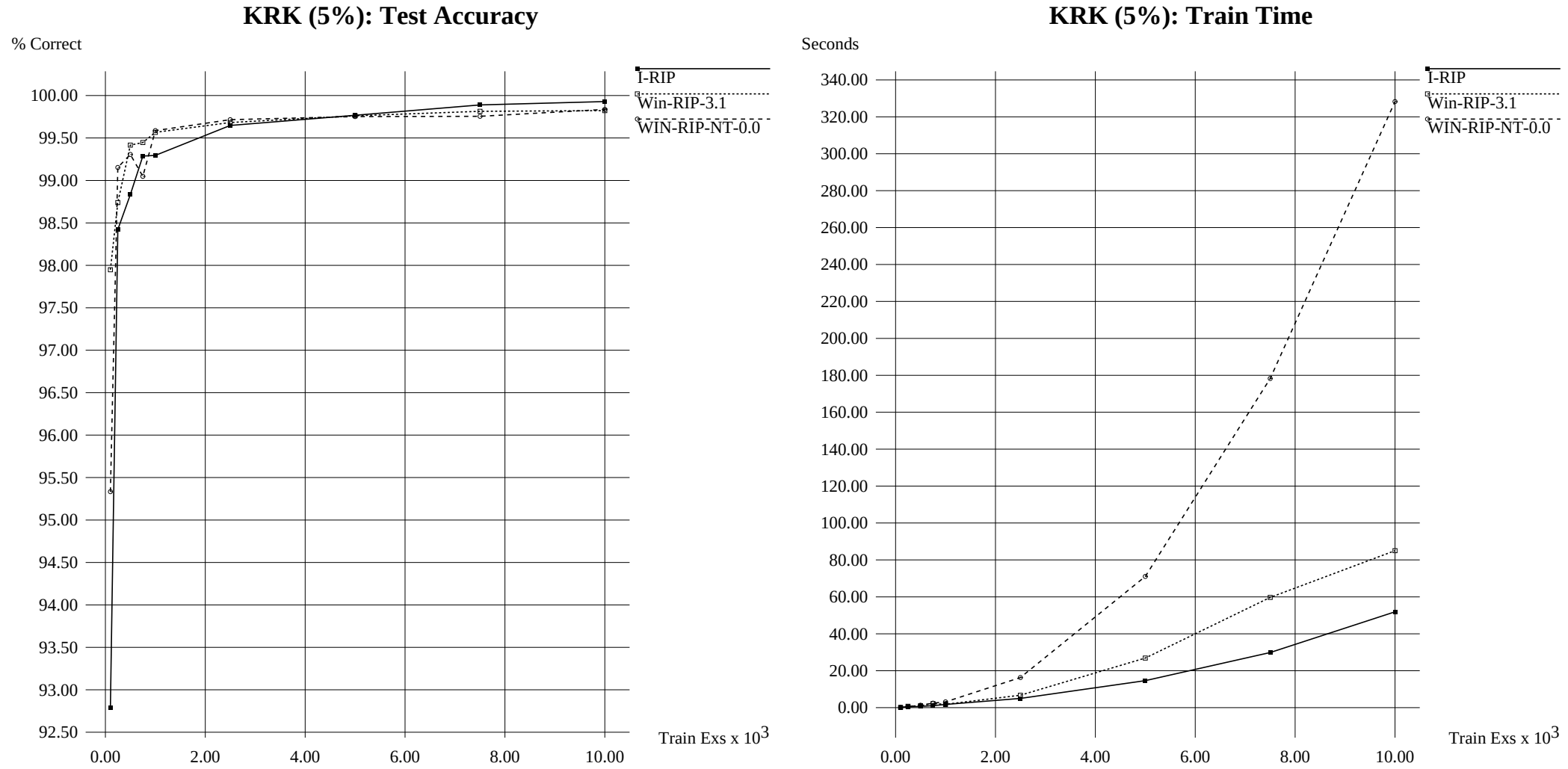

Figure 11: Results of the noise-free windowing versions for the KRK domain with 5% artificial noise.

graph for the *Mushroom* domain is shown in previous work Fürnkranz (1997d). So we can conclude that the changes in the algorithm did some harm, but the algorithm still performs reasonably well in noise-free domains.

Figure 11 shows the performance of three of these algorithms (we did not test Win-RIP-95) for datasets with 5% artificial noise. Here, I-RIP is clearly the fastest algorithm, a windowed version takes about twice as long. Although this is only a moderate level of noise, integrative windowing with Win-RIP-NT-0.0 is already unacceptably slow because it successively adds all examples into the window as we have discussed in section 5.2.1.

The interesting question, of course, is what happens if we increase the $\alpha$ parameter, which supposedly should be able to deal with the noise in the data? To this end, we have performed a series of experiments with varying noise and varying values for the $\alpha$ parameters. Our basic finding is that the algorithm is very sensitive to the choice of the $\alpha$-parameter. The first thing to notice is that there is an inverse relationship between the $\alpha$-value and the run-time of the algorithm. This is not surprising because higher values of $\alpha$ will make it easier for rules to be added to the final theory. Thus, the algorithm is likely to terminate earlier. On the other hand, we also notice a correlation between predictive accuracy and the value of $\alpha$. The reason for this is that lower $\alpha$-values impose more stringent restrictions on the quality of the accepted rules. Again, this can be explained with the fact that higher $\alpha$ values are likely to admit less accurate rules (see section 5.2.1).

Therefore, what we hope to find, is an "optimal" range for the $\alpha$-parameter for which Win-RIP-NT-$\alpha$ outperforms I-RIP in terms of run-time while maintaining about the same performance in terms of accuracy. For example, Figure 12 shows the results of I-RIP and Win-RIP-NT with a setting of $\alpha = 0.5$ in the KRK domain with 20% noise added. Both perform about the same in terms of accuracy, but there is some gain in terms of run-time.

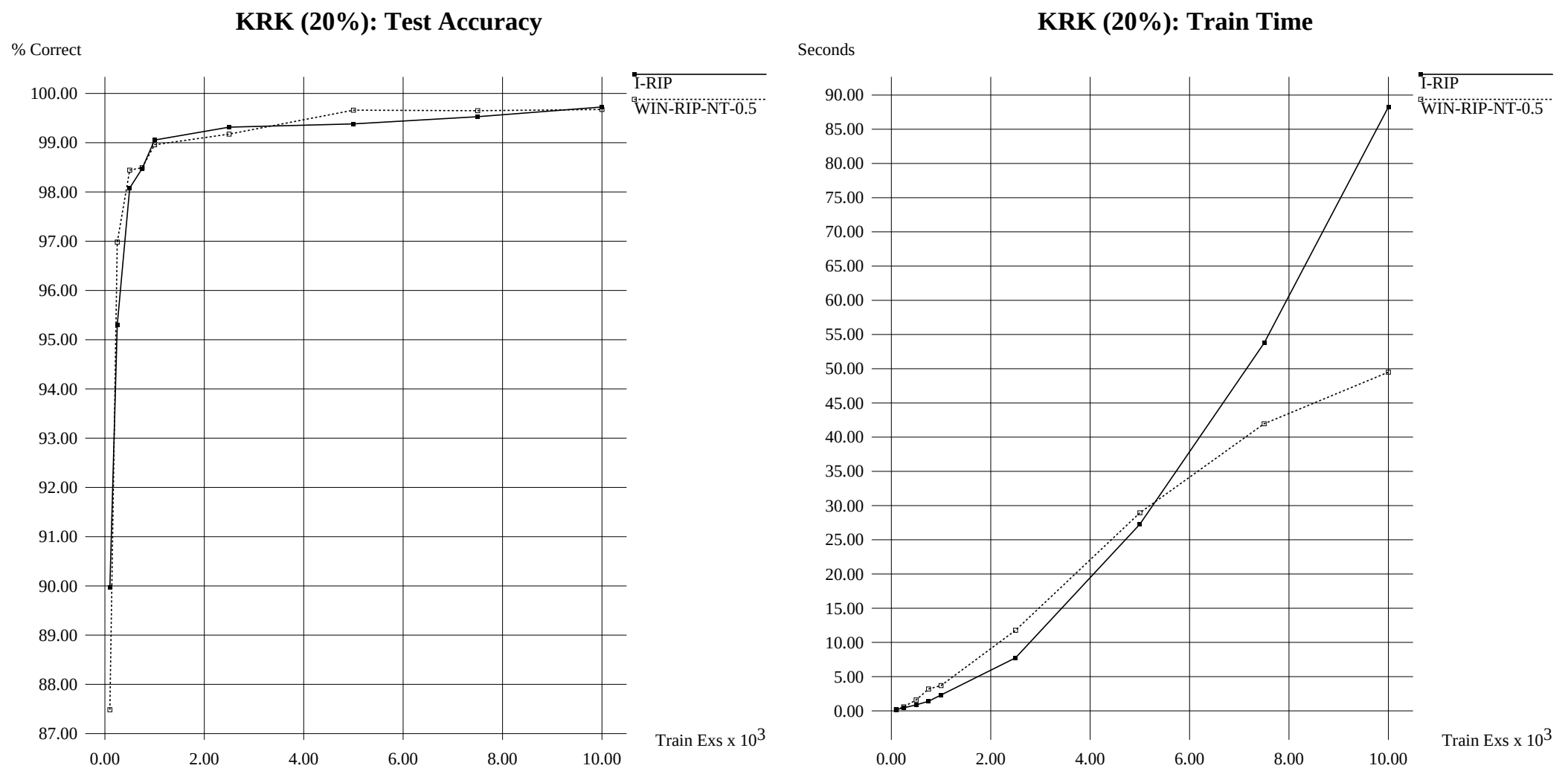

Figure 12: Results of WIN-RIP-NT with a setting of $\alpha = 0.5$ for the KRK domain with 20% artificial noise.

Although it is not as remarkable as in noise-free domains, there is a clear difference in the growth function: I-RIP's run-time grows faster with increasing sample set sizes, while WIN-RIP-NT's growth function appears to be sub-linear. Unfortunately, the performance of the algorithm seems to be quite sensitive to the choice of $\alpha$, so that we cannot give a general recommendation for a setting for $\alpha$.[16] Figure 13 shows the graph for three different settings of the $\alpha$-parameter for varying levels of noise. In terms of run-time, it is obvious that higher values of $\alpha$ produce lower run-times. In terms of accuracy, $\alpha = 0.5$ seems to be a consistently good choice. However, for example, at a noise level of 30% the version with $\alpha = 1.0$ is clearly preferable when considering the performance in both dimensions. With further increasing noise levels, all algorithms become prohibitively expensive, but are able to achieve surprisingly good results in terms of accuracy. From this and similar experiments, we conclude that there seems to be some correlation between an optimal choice of the $\alpha$-parameter and the noise-level in the data.

We have also tested WIN-RIP-NT on a discretized, 2-class version of Quinlan's 9172 example thyroid diseases database, where we achieved improvements in terms of both, run-time and accuracy with the windowing algorithms. These experiments are discussed in previous work (Fürnkranz, 1997d). However, later experiments showed that this domain exhibited the same kind of behavior as the *Shuttle* domain, shown in Figure 7, i.e., most versions of WIN-RIP-NT consistently outperformed I-RIP in terms of accuracy when learning on the minority class, while this picture reversed when learning on the majority class. Run-time gains could be observed in both cases. However, these results can only be regarded as preliminary. In order to answer the question whether our results in the KRK domain with

16. In a new domain, we would advice to try $\alpha = 1.0$ and $\alpha = 0.5$ in that order.

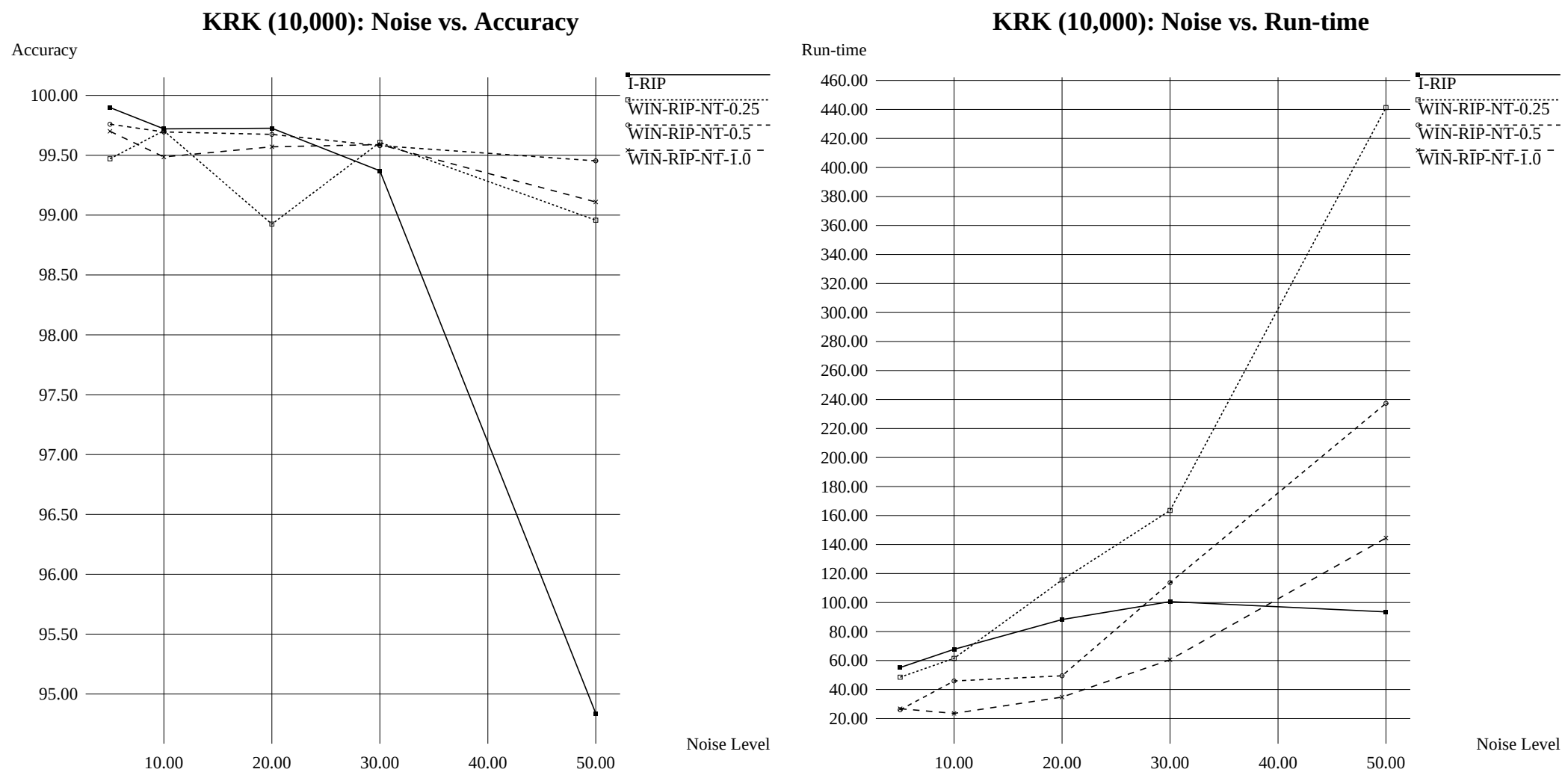

Figure 13: Results of I-RIP and three versions of Win-RIP-NT ($\alpha = 0.25, 0.5, 1.0$) for the 10,000 examples KRK domain with varying levels of artificial noise.

artificial class noise are predictive for the algorithm's performance on real-world problems, a more elaborate study must be performed.

In summary, we have discussed why noisy data appear to be a problem for windowing algorithms and have argued that three basic problems have to be addressed in order to adapt windowing to noisy domains. We have also shown that, with three particular choices for these procedures, it is in fact possible to achieve run-time gains without sacrificing accuracy. However, we do not claim that these three particular choices are optimal and are convinced that better choices are possible and should lead to a more robust learning system.

## 6. Related Work

Windowing techniques are related to several other research areas in the field of Machine Learning, including sub-sampling, active learning, and techniques for complexity reduction such as feature subset selection. In this section, we will briefly discuss some of these related techniques. For a slightly different angle on many of the following topics, the reader may also wish to consult the work by Blum and Langley (1997).

### 6.1 Sub-Sampling Techniques

Several authors have discussed approaches that use sub-sampling algorithms different from windowing. For decision tree algorithms it has been proposed to use dynamic sub-sampling at each node in order to determine the optimal test. This idea has been originally proposed, but not evaluated by Breiman et al. (1984). It was further explored in Catlett's work on *peepholing* (Catlett, 1992), which is a sophisticated procedure for using sub-sampling to eliminate unpromising attributes and thresholds from consideration.

John and Langley (1996) discuss a different approach that successively expands the learning window. ELC adds randomly selected examples to the window until an extrapolation of the learning curve (accuracy over window size) does no longer promise significant gains for further enlargements of the window. In terms of run-time, the authors note that this technique can in general only gain efficiency for incremental learning algorithms.

*Partitioning*, proposed by Domingos (1996b, 1996a), splits the example space into segments of equal size and combines the rules learned on each partition. This technique produced promising results in noisy domains, but has substantially decreased learning accuracy in non-noisy domains. Besides, the technique seems to be tailored to a specific learning algorithm and not applicable to a wider group of rule learning algorithms (Domingos, 1996a).

Sub-sampling techniques have recently been frequently used for increasing the accuracy of a classifier. The basic idea is to train classifiers on multiple subsamples of the data and combine their predictions, usually by voting. Such approaches include *bagging* (Breiman, 1996b), where all examples are sub-sampled with equal probabilities, and *boosting* (Drucker, Schapire, & Simard, 1993; Freund & Schapire, 1996) — also known as *arcing* (Breiman, 1996a) — where examples that have been misclassified in previous iterations are more likely to be selected in the next iterations. Interestingly, Breiman (1996b) has noted that these techniques rely on unstable base learners, while we conjecture that the better performance of windowing with rule learning algorithms is due to more stability (section 3.3). While the main focus of these works is to improve the accuracy of a given learning algorithm, it would be interesting to evaluate bagging and boosting techniques along their run-time and memory requirements as well. For example, Breiman (1996c) discusses arcing algorithms for datasets where limited memory requires the use of sub-sampling.

In Knowledge Discovery for Databases, sub-sampling techniques have been investigated for the discovery of association rules. Toivonen (1996) describes a straight-forward approach, where association rules are discovered from a subsample of the data and their validity is checked on the complete dataset. Kivinen and Mannila give theoretical bounds for the sample size that is required for establishing (with a given maximum error probability) the truth of association rules (Kivinen & Mannila, 1994) or functional dependencies (Kivinen & Mannila, 1995) that have been discovered from the sample.

### 6.2 Active Learning

Windowing techniques are also closely related to the field of *active learning*. According to the term's original definition (within the field of Machine Learning), active learning includes "any form of learning in which the learning program has some control over the inputs it trains on." (Cohn, Atlas, & Ladner, 1994). While this definition would be broad enough to include windowing techniques, subsequent work in this area has mostly concentrated on the use of *membership queries*, i.e., on giving the learner the means to query the classification of examples of its own choice instead of providing it with a fixed set of labeled data.

Such approaches are based on a new motivation for studying sub-sampling techniques (in addition to the three motivations listed in the Introduction to this paper):

**Expensive Labeling:** In many domains, it is very expensive to obtain pre-classified training examples, while unlabeled examples are cheaply available. Sub-sampling tech-niques can help to solve this problem by focussing on minimizing the number of examples that have to be labeled in order to learn a satisfying theory.

The prevalent example of such a domain is the World-Wide Web, which provides an abundance of training pages for text categorization problems. However, significant effort is required to assign semantic categories to these pages. Not surprisingly, much of the recent work in active learning has concentrated on text categorization problems.

Closely related to windowing is *uncertainty sampling* (Lewis & Gale, 1994; Lewis & Catlett, 1994). The difference is that the window is not adjusted on the basis of misclassified examples, but on the basis of the learner's confidence in its own predictions. The examples that are classified with the least confidence will be added to the training set in the next iteration (after obtaining their class labels).

However, not all learning algorithms are able to attach uncertainty estimates to their predictions. Besides, using uncertainty estimates from a single learning algorithm may be problematic in some cases (Cohn et al., 1994). Thus, it was suggested to use a committee of classifiers and measure the uncertainty in the predictions by the degree of disagreement among the classifiers. The *selective sampling* technique proposed by Cohn et al. (1994) is one such technique, which is based on a theoretical framework that uses the entire version space of consistent theories as a committee. Another version of this approach is the *query by committee* algorithm (Seung, Opper, & Sompolinsky, 1992; Freund, Seung, Shamir, & Tishby, 1997), which uses a probability distribution over hypotheses to randomly select two consistent hypotheses for classifying a new example. If their predictions differ, the algorithm asks for the true label of the example and adds it to the training set. *Committee-based sampling* (Dagan & Engelson, 1995) is an adaptation of this idea to probabilistic classifiers. Liere and Tadepalli (1997) compare several ways of combining the predictions of a committee of Winnow-based learners on a text categorization task

Obviously, the above-mentioned approaches can also be applied to situations where a large amount of labeled data is available. An investigation of the suitability of these techniques for increasing the efficiency of learning algorithms and a comparison to our solutions for this problem is left for future work. However, it should be noted that these techniques are not susceptible to the problem of noise in the form we discussed in section 5.1, because they would simply ignore the class labels during the subsampling process.

It is less obvious that windowing techniques might also be useful in the presence of large amounts of unlabeled data. However, several authors have recently started to investigate the idea that, instead of submitting the predictions that the learner is most uncertain about to a teacher for labelling, the learner might autonomously label the predictions it is most certain about and add them to the training set. Proposals include the use of a committee of learners for *co-training* (Blum & Mitchell, 1998) or techniques based on the Expectation Maximization (EM) algorithm (Nigam, McCallum, Thrun, & Mitchell, 1998), possibly coupled with an active learning procedure (McCallum & Nigam, 1998). If research in this direction is pushed further, the number of labeled examples available to a learning algorithm might be greatly enlarged at the expense of some incorrectly labeled examples. Noise-tolerant windowing algorithms may turn out to be an appropriate choice for such data sets.

```
procedure WINDOWING(Algorithm,LP)

RedLP = INITIALIZEREDUCTION(LP)
loop
    Theory = CALL(Algorithm,RedLP)
    Q = EVALUATE(Theory,LP)
    if STOPPINGCRITERION(Q,LP,RedLP)
        return(Theory)
    else
        RedLP = EXPANDREDUCTION(Q,LP,RedLP)
```

Figure 14: A general view of windowing.

### 6.3 Complexity Reduction

Windowing can be viewed as special case of a wider variety of optimization techniques that try to reduce the complexity of a learning problem by identifying appropriate low-complexity versions of the problem. The complexity of a learning problem mostly depends on the number of training examples and the size of the searched hypothesis space.[17] Figure 14 shows an abstraction of the windowing algorithm. It starts by initializing the learning problem with a reduced learning problem (e.g., with a subsample of the examples), then applies the learning algorithm to this reduced problem and analyzes the resulting theory with respect to the original problem. Unless some stopping criterion specifies that the quality of learned theory is already sufficient (e.g., if no exceptions could be found on the complete data set), the reduced learning problem will be expanded to incorporate more information (e.g., by adding all misclassified examples) and a new theory is induced.

Note that this abstract framework also describes other approaches for reducing the complexity of a learning problem, such as hypothesis space reduction. As an example think of an algorithm that attempts to learn a theory in a simple hypothesis space first and only switches to more complex hypothesis spaces if the result in the simple space in unsatisfactory. For example, many *Inductive Logic Programming* systems provide some explicit control over the complexity of their hypothesis space, which might be controlled with an instantiation of the generalized windowing algorithm. Such an approach has in fact been realized in CLINT (De Raedt & Bruynooghe, 1990), which has a predefined set of hypothesis spaces with increasing complexity, and is able to switch to a more expressive hypothesis language if it is not able to find a satisfactory theory in its current search space. Similar approaches could also be imagined for other ILP algorithms. For example, in FOIL (Quinlan, 1990), one could systematically vary certain parameters that influence the complexity of the hypothesis space, like the number of new variables that can be introduced in the body of a clause or the maximum length of a clause, in order to define increasingly complex hypothesis spaces. This procedure could be automatized in a way similar to the one described by Kohavi and John (1995). The crucial point is how to efficiently evaluate that no progress can be made

17. In representation languages that extend flat feature vectors, such as first-order logic, the complexity of a learning problem also depends crucially on the average cost of matching an instance with a rule. Sebag and Rouveirol (1997) demonstrate a technique for reducing these potentially exponential costs via sub-sampling.

by shifting to a more complex hypothesis space. In the propositional case, *forward selection* approaches to feature subset selection, i.e., algorithms that select the best subset of features by adding one feature at a time to an initially empty set of features, can be viewed in this framework (Caruana & Freitag, 1994; Kohavi & John, 1997).

All of the above-mentioned approaches may be viewed as a particular type of *bias shift* operator (Utgoff, 1986; desJardins & Gordon, 1995) focussing on shifts to computationally more expensive inductive biases. Turney (1996) investigates this in more detail, but suggests that — in order to maximize accuracy — one should start with a weak bias and gradually shift to stronger biases. Our results suggest the opposite strategy if efficiency is the main concern. This is consistent with results in comparing forward and backward feature subset selection (Kohavi & John, 1997).

We believe that thinking in this framework may lead to more general approaches for reducing the complexity of a learning problem, which aim at reducing both hypothesis and example space at the same time. As an example consider the *peepholing* technique introduced by Catlett (1991b), where sub-sampling is used to reliably eliminate unpromising candidate conditions from the hypothesis space.

## 7. Future Work

There are several ways how future work can be based on the presented results. Clearly, a deeper exploration of the effects of the parameter settings of our algorithms is needed. In particular, a better understanding of the $\alpha$-parameter in the noise handling heuristics is necessary for practical applications of the algorithm. Alternative solution attempts for the three problems outlined in section 5.2 should also be investigated, thus maybe entirely eliminating the need for this parameter. As discussed in section 3.4, the applicability of windowing algorithms crucially depends on the presence of some redundancy in the training set. Thus, better methods for characterizing redundant domains are a rewarding topic for further research. The efficiency of the presented algorithms could maybe be further improved by trying to specialize over-general rules, instead of entirely removing them from the current theory and relying on the next windowing iteration to find a better rule. Ideas from theory revision might turn out to be useful in this context.

While our major concern was to demonstrate that significant increases in run-time are possible, it might be promising to further investigate the use of windowing techniques for increasing predictive accuracy or for increasing the effectiveness for learning with limited memory resources. For the former problem, techniques for combining the results of multiple runs of windowing with different random seeds, as implemented in C4.5 (Quinlan, 1993), are quite similar to bagging and boosting techniques and could produce similar results. For learning in the presence of memory limitations, we have found that in noise-free domains, the curve for the total number of examples submitted to the learning algorithms flattens at the point where 100% training accuracy is reached. Thus the windowing algorithms need asymptotically less memory than the learning algorithms per se, if one assumes that the testing of learned theories can be performed on disk. However, no guarantees can be made whether the use of windowing will allow a given learning problem to be solved without exceeding a given memory bound. A closer investigation of these issues would also be a rewarding topic for further research.

Currently, the applicability of our algorithms is limited to propositional 2-class problems. A straightforward adaption to multi-class problems can be performed in several ways (Clark & Boswell, 1991; Ali & Pazzani, 1993; Dietterich & Bakiri, 1995). A different approach might adapt the algorithm for learning ordered rule sets as in CN2 (Clark & Niblett, 1989). For example, one could use a separate windowing process for each rule, so that the rules that are successively added to the final theory can be of different classes. Orthogonally, an extension of windowing for first-order learning techniques is another challenging and rewarding task. Note that the sub-sampling problem in inductive logic programming is considerably harder than for propositional learning. For example, first-order learning systems often allow an implicit definition of training examples (Muggleton, 1995) or learn from positive examples only by making some form of closed-world assumption (Quinlan, 1990), which prevents the straight-forward use of sub-sampling. Furthermore, the examples in training sets for ILP algorithms can depend on each other. For example, learning a recursive concept often requires that all instantiations of the target predicate that are used in the derivation of a positive example are present in the training data, a property which is not likely to hold in the presence of sub-sampling.

## 8. Conclusion

In this paper, we have presented a re-evaluation for windowing with separate-and-conquer rule learning algorithms. For this type of algorithm, significant gains in computational efficiency are possible without a loss in terms of predictive accuracy. We explain this with the fact that separate-and-conquer rule learning algorithms learn each rule independently, whereas an attribute chosen by a divide-and-conquer decision tree learning algorithm will be part of all rules that are represented by the subtree below this attribute. Based on this finding, we have further demonstrated a more flexible technique for integrating windowing into rule learning algorithms. Good rules are immediately added to the final theory and the covered examples are removed from the window. This avoids re-learning these rules in all subsequent iterations of the windowing process thus reducing the complexity of the learning problem. While most of our results have been obtained in noise-free domains, we believe that the idea of integrative windowing can be generalized for attacking the problem of noise in windowing. To that end, we have outlined three basic problems that have to be solved. A first implementation of straightforward solutions to these problems has achieved promising results in a simple domain with artificial noise.

## Acknowledgements

This research is sponsored by the Austrian Fonds zur Förderung der Wissenschaftlichen Forschung (FWF) under grant number J01443-INF (Schrödinger-Stipendium). I would like to thank Hendrik Blockeel, William Cohen, Tom Mitchell, Bernhard Pfahringer, Michèle Sebag, Ashwin Srinivasan, Gerhard Widmer, and the anonymous reviewers of this and previous versions of the paper for helpful suggestions, discussions and pointers to relevant literature. Thanks are also due to Ray Mooney for making his Common LISP Machine Learning library available and to the maintainers of the UCI Machine Learning repository.